\documentclass[twocolumn,twoside]{IEEEtran}
\IEEEoverridecommandlockouts 

\usepackage{graphicx} 
\usepackage{amsmath}
\usepackage{amssymb}
\usepackage{caption}
\usepackage{subcaption}
\usepackage{multirow}
\usepackage{multicol}
\usepackage{booktabs} 
\usepackage{hyperref}
\newcommand*{\email}[1]{\href{mailto:#1}{\nolinkurl{#1}} } 
\usepackage[table]{xcolor}
\usepackage{array,multirow,graphicx}

\graphicspath{ {figs/} }
\usepackage{tikz}
\usepackage[normalem]{ulem}

\usepackage{lipsum}

\usepackage{tikz} 
\usetikzlibrary{matrix}

\usepackage{amsmath}

\begin{document}


%

\title{A Human-Inspired Thumb-Index Robotic Hand with Strain Gauges Embedded in Soft Joints}

\author{Jonas~Papenbrock$^{\dagger,1}$, Shubhan~Patni$^{\dagger,2}$, Tomaso~Lisini~Baldi$^{1}$, Michele~Guerri$^{1}$, Elia~Landi$^{1}$, Ada~Fort$^{\ast,1}$, and~Matej~Hoffmann$^{\ast,2}$   
\thanks{$^\dagger$Shared first authorship. $\ast$Equal senior contribution.}
\thanks{$^1$ Department of Information Engineering and Mathematics - University of Siena.}
\thanks{$^2$Department of Cybernetics, Faculty of Electrical Engineering, Czech Technical University in Prague \email{matej.hoffmann@fel.cvut.cz} .}
\thanks{The contribution from the University of Siena are co-funded by the European Union and the Italian Ministry of Research, under the complementary actions to the NRRP ``Fit4MedRob-Fit for Medical Robotics'' under Grant PNC0000007 (CUP B53C22006930001).}
\thanks{The contribution from the Czech Technical University in Prague is co-funded by the European Union under the ROBOPROX (Robotics and advanced industrial production) project (reg. no. CZ.02.01.01/00/22\_008/0004590). S.P. was additionally supported by the Grant Agency of the Czech Technical University in Prague, grant No. SGS26/075/OHK3/1T/13 .}}

\ifCLASSOPTIONpeerreview
    \IEEEpeerreviewmaketitle
\else
    \maketitle
\fi

\begin{abstract}

Human hand grasp adaptation depends mainly on the synergy between physical structure and biological feedback. Inspired by this biomechanical principle, the Safe Thumb-Index Robotic (STIR) Hand was developed as a minimal, lightweight, and low-cost two-digit prototype featuring an asymmetric thumb-index configuration. By pairing an underactuated, tendon-driven mechanical design with flexible strain gauges embedded into silicone-encapsulated soft joints, the system achieves passive grasp adaptation while establishing both internal proprioception and external perception. Unsupervised analysis was carried out on a dataset of the STIR hand grasping 20 different objects, along with an object classification task and an ablation study to highlight the contribution of the soft joint sensors. The object classification task discriminated object size, shape, and material stiffness with a high classification accuracy. In contrast to traditional industrial grippers and robotic hands, the STIR Hand demonstrates that sensorized compliant joints significantly improve overall sensitivity and ensure safe grasping, while remaining independent of additional fingertip tactile elements or external vision systems. Finally, a comparison to similar devices grasping identical objects validates the utility of the STIR Hand.

\end{abstract}


\section{Introduction}
\label{sec:introduction}

Underactuated and soft robots are appealing for their natural compliance, safety, and their potential to exploit intrinsic dynamics to simplify control \cite{chen2025bioinspired, Pfeifer2007self}. The combination of flexible material properties and underactuated kinematics enables the system to passively adapt its shape to irregular geometries during physical interaction \cite{laschi2014soft}. This makes hybrid soft-rigid designs highly applicable to assistive devices, prosthetic systems, and delicate manufacturing tasks \cite{Rus2015design}.

Monitoring and controlling compliant hands remains a major challenge. Low-cost soft hands rarely feature an accurate proprioceptive sense, frequently relying on indirect metrics such as motor currents, actuator outputs, and tactile sensors to estimate finger positions. To track the structural changes of the device, existing approaches have integrated highly flexible strain sensors or liquid-phase elastomeric channels into soft structures \cite{Amjadi2014highly}. However, these sensing technologies frequently suffer from signal noise, limited lifecycles, weak material bonding, and complex manufacturing processes \cite{wang2018toward}. In contrast, metal thin-film or foil strain gauges offer predictable and linear behavior, but do not tolerate large deformations. Through a novel joint design that embeds these strain gauges in a specific position, the STIR Hand utilizes precise output within a highly flexible structure.

To resolve these coupled mechanical and sensing constraints, the Safe Thumb-Index Robotic (STIR) Hand introduces proprioception to a minimal, two-digit design. This prototype features an asymmetric thumb-index layout and soft sensorized joints. Building upon recent work in embedded flex sensing within wearable assistive devices \cite{salvietti2018robotic, landi2025sensorization}, the STIR Hand pairs an underactuated, tendon-driven mechanical architecture with foil strain gauges embedded inside silicone-encapsulated soft joints. By strategically positioning the strain gauges near the neutral axis of the bending joint, measurement range limitations are mitigated, enabling effective operation within a small deformation window while achieving a highly stable, precise, and linear sensor response. This design provides localized deformation data regarding structural compliance, which is combined with global motor-current feedback to capture high-resolution interaction profiles without requiring oversized fingertip tactile systems or external vision devices.

This study provides two key contributions: 
\begin{enumerate}
    \item \textbf{Hardware Innovation:} The design and systematic engineering of a low-cost, underactuated hardware platform are presented, featuring robust mechanical coupling and foil strain gauges embedded into the soft joints. The novel development of these strain gauges is their implementation near the neutral bending axis to remain within the strain limits of the sensors. 

    \item \textbf{Empirical Dataset Creation:} A comprehensive data collection framework is established to evaluate diverse physical object interactions across size, geometry, and material stiffness.
\end{enumerate}

\ifCLASSOPTIONpeerreview
    All of our data, classification models, and analysis is made open source at \url{https://anonymous.4open.science/r/STIR-Hand-EEFA}.
\else
    All of our data, classification models, and analysis is made open source at \url{https://github.com/ShubhanPatni/STIR-Hand}.
\fi

\section{Related work}
\label{sec:rel_work}
Bio-inspired designs that supplement active control with passive autonomy, much like the heart in the human body, are a promising solution for robotic integration into assistive applications~\cite{vanLaake2024}. Underactuated hands and grippers utilize fewer actuators and using soft parts which provide morphological adaptation simplifies control without requiring extensive computational resources~\cite{JungeHughes2025}. Bio-inspiration has driven robotic solutions towards hybrid designs where rigid segments provide structural stability and soft segments facilitate safe, energy-efficient environmental contact~\cite{Lyu2025Humanoid}.

\subsection{Underactuated hands}
\label{subsec:underactuated_hands}

An exemplary case of how underactuated hands can operate efficiently is the tactile soft hand described in~\cite{SoftHand2014}, which demonstrates that a 3D-printed, two-actuator gripper can achieve versatile manipulation alongside robust grasping adaptability. This research shows that tendon-driven grippers could maintain flexible handling while reducing the total number of actuators. Similarly, the gripper in~\cite{Dong2025ModelQII} utilizes flexible joints between rigid segments to switch between power grasps and precise pinch manipulation through passive mechanisms with an additional position-locking system, but this increases fabrication and maintenance difficulty. Another underactuated design described in~\cite{Cloth2025} alters grasping strategies by switching from point to linear grasping to handle highly deformable materials. This device combines a low degree-of-freedom architecture with a suction system, but the pneumatic subsystem increases hardware complexity and reduces suitability for wearable devices. To achieve dexterous grasping without extra actuators, a three-finger gripper in~\cite{Passive2025} uses passive joints reconfigurable via external forces, but integrating feedback sensors near these joints increases finger size. Taking a different approach, the modular gripper in~\cite{DuverneyRauter2026} estimates grasping forces through finger deformations via an integrated camera. While it handles both rigid and soft objects using visual feedback, internal cameras are sensitive to variable lighting and require frequent calibration. Finally, a humanoid hand with omnidirectional soft bending sensors utilizing segmented optical fibers and trichromatic LEDs was presented in~\cite{Zhong2026SoftSensor}. This hybrid rigid-flexible finger structure enables posture perception for delicate tasks, and by placing motors near the joints, it reduces wiring and system complexity, but motor size limitations reduce its capability for handling heavier objects. 

Overall, these experimental devices confirm that underactuated hands provide adaptable solutions for robust grasping, exploiting passive conformation to the object shape. However, various limitations regarding precision control, mechanical complexity, and fabrication persist across different designs. The Safe Thumb-Index Robotic (STIR) Hand overcomes these challenges by combining the core advantages of an underactuated system, specifically offering low mechanical complexity, reduced system weight, and low production costs. Moreover, the STIR Hand utilizes sensorized elastic joints to provide passively robust, adaptive mechanics while simultaneously enabling precise kinesthetic perception.

\begin{figure*}[!h]
    \centering
    \includegraphics[width=0.95\textwidth]{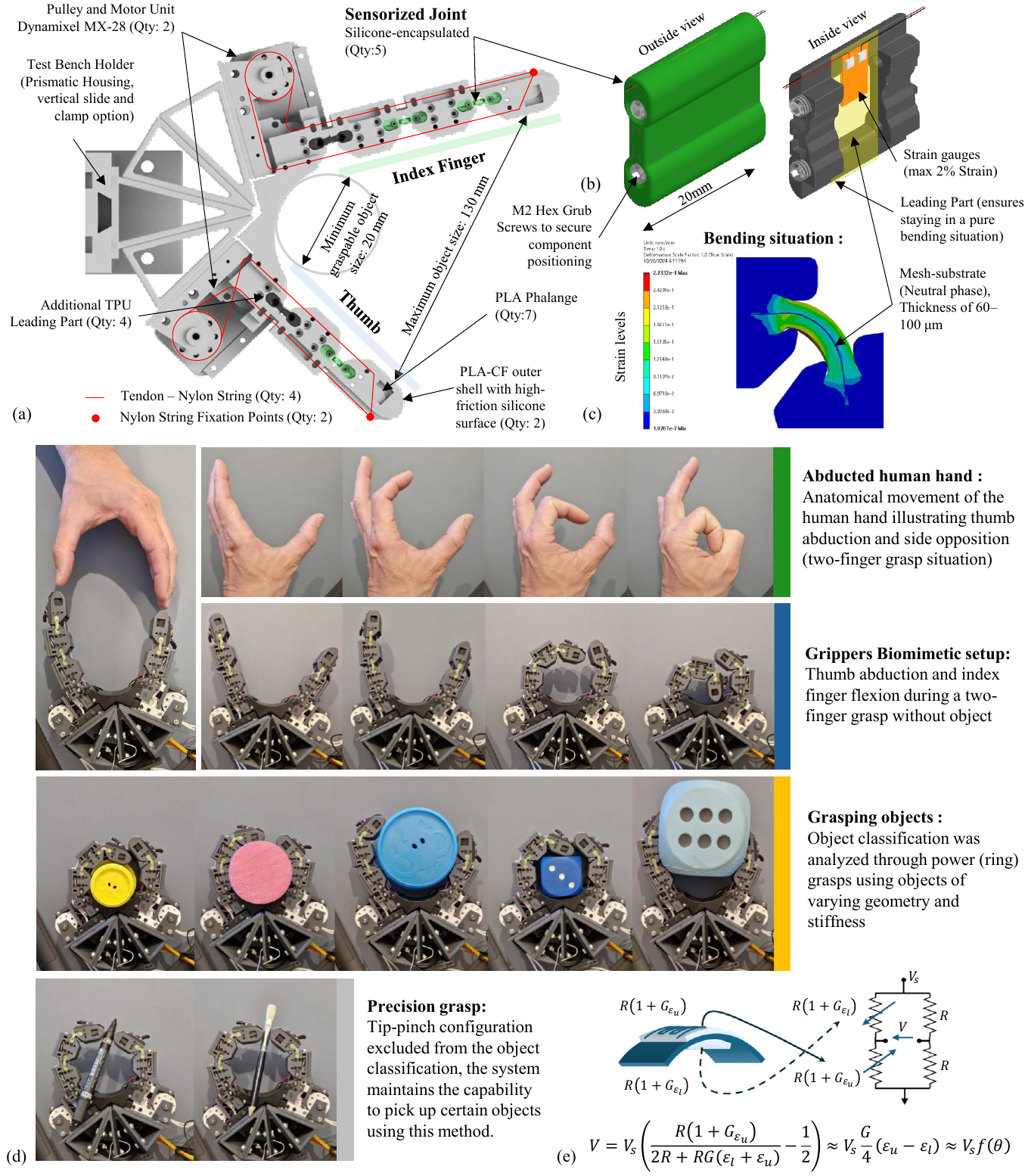}
    \caption{
Overview of the modular STIR-Hand: (a) Design assembly of the thumb and index finger showing grasping dimensions for various objects; (b) outside and inside views of the Sensorized Joint  \cite{landi2025sensor} featuring its strain gauges, mesh substrate, Leading Part  \cite{landi2025sensor}, hex grub screws, and wiring; (c) visualization of the neutral axis (blue regions in the center of the joint) under bending deformation, illustrating strain levels when a mesh substrate is integrated into the silicone body; (d) various grasping modes of the dual-finger STIR-Hand in comparison with a human hand; (e) half-bridge sensor configuration and the linear relationship between differential strain and joint angle as defined by the equation.
    }
    \label{fig:STIR-Hand_Design}
\end{figure*}

\subsection{Haptic sensing and object recognition}
\label{subsec:haptic_sensing}
Major advances in haptic sensing technology for robotics have been made in the form of fingertip tactile sensors operating on different sensing principles, for example piezoelectric sensing~\cite{wettels2008biomimetic}, light sensing~\cite{khamis2019novel}, magnetic field sensing~\cite{tomo2017covering}, and visual input~\cite{lambeta2020digit}. However, integrating feedback and sensing from other parts of the hand, in particular the digit phalanges, remains underexplored. These proprioceptive components have been shown to increase grasp stability and also improve task success rates for dynamic tasks like manipulation~\cite{zhou2026tactile}. Further, different robot morphologies, gripper-object pairs, sensor configurations, and exploratory actions in object classification scenarios, have been shown to influence the accuracy of different models in object classification tasks~\cite{pliska2024single}. We integrate bending sensors into the articulated joints of our gripper fingers. These sensors provide state checks of the gripper, while also acting as feedback for sensing external torques and resistances generated while grasping or manipulating objects. 

Object recognition and classification have become standard tasks for evaluating the performance of novel tactile sensors for robotics~\cite{zhang2025visual} in real-world applications. The tactile sensors are mounted on a robotic end-effector and made to sense different objects through an interaction sequence. Then, the collected data is analyzed through statistical or learning-based methods and classifiers capable of sorting the objects based on the preferred object properties are developed. Beyond classification accuracy, this task paradigm allows for the evaluation of the sensor's capabilities. By systematically varying chosen target properties like surface texture, surface friction, hardness and elasticity, correlations can be found about how the sensor's measurements represent these variations. Systematic variation of object properties can be seen in the object set designed by Dutta et al.~\cite{dutta2026embodied}, which used 3D printing to create parametric variations in the surface texture using spatial frequency and amplitude, and used different base materials to create 'soft', 'semi-soft', and 'hard' categories for object stiffness. We have devised our own object sets (Section~\ref{subsec:obj_datasets}) for the purpose of similar systematic exploration.

\section{Material and Methods}
\label{sec:Materials_Methods}

\subsection{Safe Thumb-Index Robotic Hand}
\label{subsec:Safe Thumb-Index Robotic Hand}
The STIR-Hand is an underactuated, safe, tendon-driven, two-finger robotic hand actuated by two independent motors as mentioned in Section ~\ref{sec:introduction}. An overview of the modular STIR-Hand and its main components is shown in Fig.~\ref{fig:STIR-Hand_Design}. In the following points, its most important characteristics are explained:

\subsubsection{Hardware Mechanics and Sensorized Joint}
\label{subsub:Hardware Mechanics and Sensorized Joint}
The rigid segments, fabricated from PLA and PLA-CF, provide geometric stability with defined constraints, while the soft joints, fabricated from silicone, nylon and TPU enable safe and adaptive grasping. Two nylon tendons per finger are routed through the rigid segments and connected to the motor unit (Dynamixel MX-28, Protocol 2.0) via pulleys arranged in clockwise and counterclockwise directions. To enforce a pure bending situation and to guarantee accurate and stable angle measurements, a  ``Leading Part'' is implemented into the soft joints. This part features a defined collapse zone near the center of rotation between the phalanges, which is dictated by the central cut-off of the Leading Part \cite{landi2025sensor}. The joint stiffness can be optimized via the Leading Part 3D-printing configurations, customizing the joints for specific tasks, from delicate handling to high-payload manipulation. 

\subsubsection{Control and Sensing Electronics}
\label{subsub:Control and Sensing Electronics}
\begin{figure}[!h]
    \centering
    \includegraphics[width=0.48\textwidth]{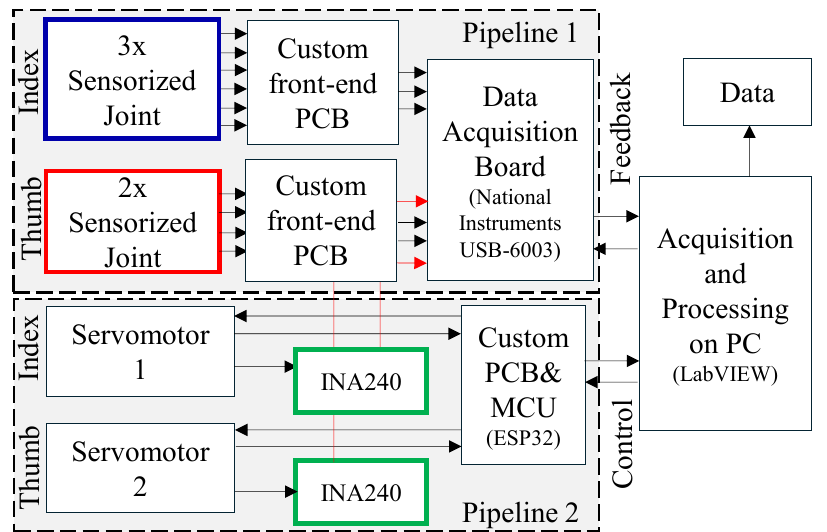}
    \caption{Block diagram of the conditioning electronics and data acquisition . The border colors highlight the feedback components as per Fig.~\ref{fig:feedback}.}
    \label{fig:Conditioning Electronics}
\end{figure}

Sensing and actuation have independent hardware to keep the two pipelines decoupled, as shown in Fig.~\ref{fig:Conditioning Electronics}. This layout ensures that both pipelines run at a synchronized frequency of 500~Hz, which is also the rate at which the trajectory is logged. The STIR Hand feedback signals are sampled by a National Instruments (NI) USB-6003 data acquisition board and processed in a LabVIEW-based software application \cite{landi2025sensor,landi2025sensorizing}. The strain gauge signals pass through a custom front-end PCB, especially designed for the STIR Hand in a half-bridge configuration (Fig.~\ref{fig:STIR-Hand_Design}e) and route to the NI USB-6003. The INA240 current-sense amplifier routes the signal to the NI USB-6003 and enables signal compensation during grasping, as detailed in Section~\ref{subsub:Sensorized Joint and Proprioception}, while the two Dynamixel MX-28 servo motors are controlled by a custom PCB featuring an ESP32 microcontroller (MCU) over the Dynamixel Protocol 2.0 bus at 1~Mbps. This modular separation keeps the actuator command loop deterministic and avoids resource contention with the strain gauge acquisition.

\subsubsection{Sensorized Joint and Proprioception}
\label{subsub:Sensorized Joint and Proprioception}
\begin{figure}[!h]
    \centering
    \vspace{-10pt}
    \includegraphics[width=0.47\textwidth]{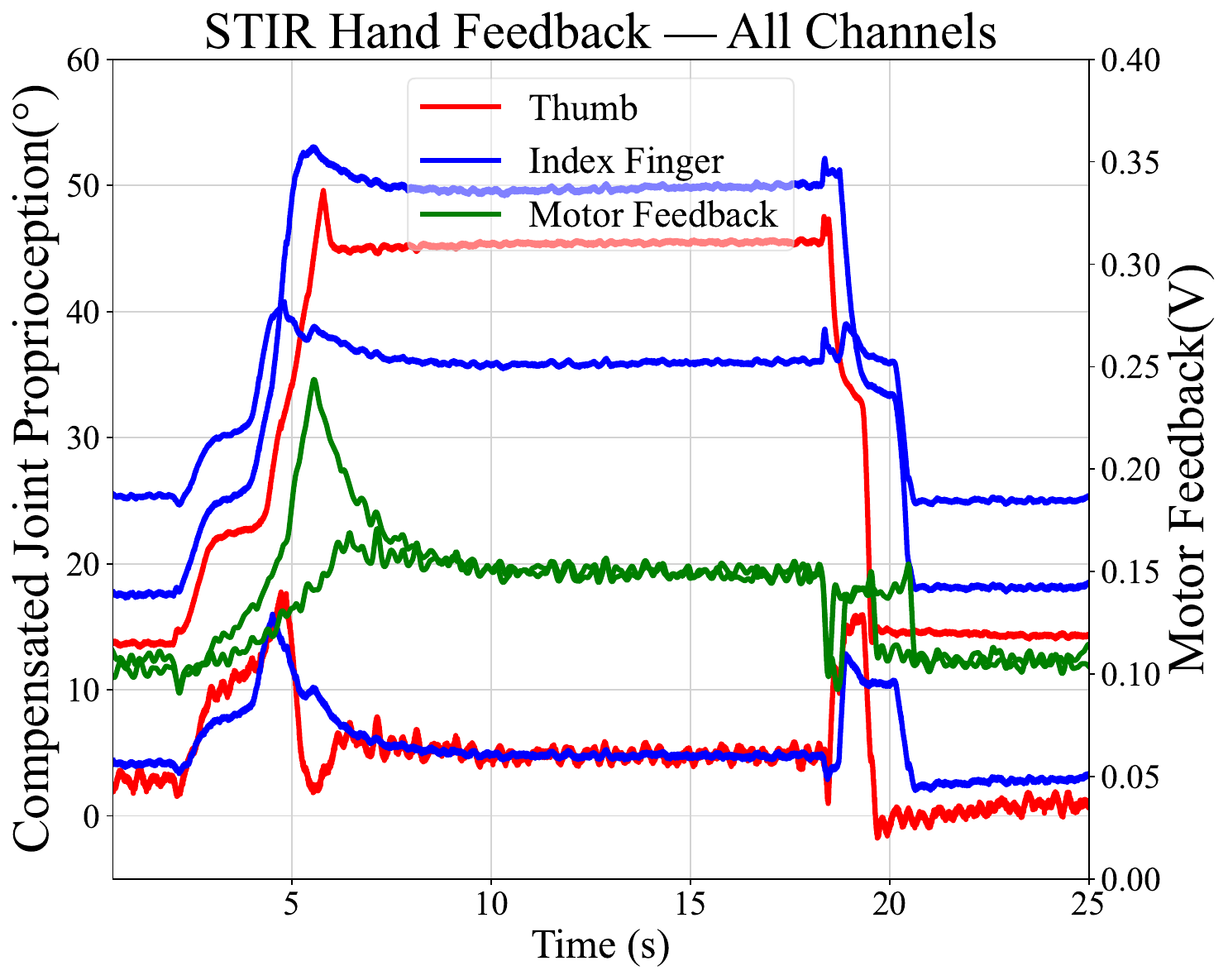}
    \caption{Example of feedback collected from the gripper, which includes 2 thumb joints, 3 index finger joints, and 2 motor voltages.}
    \label{fig:feedback}
\end{figure}

Each joint was calibrated by sampling multiple bending positions to obtain the voltage--angle correlation. The resulting correlation exhibited a linear relationship, reducing computational complexity \cite{landi2025sensorizing}. The strain gauges (BF350-3AA) were bonded to a flexible nylon mesh substrate (62T monofilament nylon mesh), as shown in Fig.~\ref{fig:STIR-Hand_Design}(b). Due to the use of a very thick mesh substrate and the forced pure bending situation by the Leading Parts, the resulting deformation remains under the limits of the strain gauge within 2\% strain, as shown in Fig.~\ref{fig:STIR-Hand_Design}(c). Proprioception in the STIR-Hand combines the embedded strain gauges with motor-current feedback from the two Dynamixel units, sampled by custom INA240-based amplifiers \cite{landi2025sensorization}. During quasi-static grasps, the motor current is used to compensate for the strain gauge readings. This compensation occurs when the finger has blocked against the object but the contact force keeps increasing. The gain Angle while the actual joint angle remains constant. This decoupling ensures that the estimated angle reflects only the true deformed values of the joint bending. The calibrated and compensated signals from the Senozorized Joint in degrees and the Motor Feedback in Voltage are shown in  Fig.~\ref{fig:feedback}.

\subsubsection{Motor Torque Estimation}
\label{subsub:Motor Torque Estimation}
The prototype uses servo motors with torque capacities exceeding the grasping requirements. Motor torque $\tau$ is estimated offline from the current sampled by the INA240 amplifier, where the hardware gain dictates $I = V_0 / 3$. Using the manufacturer's stall data for the MX-28 at $12\text{ V}$ ($\tau_{\text{stall}} = 2.5\text{ Nm}$ at $I_{\text{stall}} = 1.4\text{ A}$, yielding $K_t = 1.786\text{ Nm/A}$) \cite{robotis_emanual}, the voltage-to-torque mapping becomes:

\begin{equation}
\tau = I \cdot K_t = \left( \frac{V_0}{3} \right) \cdot \left( \frac{2.5}{1.4} \right) \approx 0.595 \cdot V_0 \, \text{[Nm]}
\end{equation}

This relationship only converts the motor effort into physical units, while the real-time controller uses the servo motor's Load parameter, as detailed in Section~\ref{subsub:Real-Time Grasp Control}

\subsubsection{Real-Time Grasp Control}
\label{subsub:Real-Time Grasp Control}
The MX-28 firmware exposes a read-only parameter named Load (register 126), which is internally inferred from the PWM duty cycle and is therefore not a calibrated measurement of torque or current \cite{robotis_emanual}. This Load parameter is nevertheless adopted as the real-time control variable because it is filtered by the firmware, available on the Dynamixel bus with negligible latency, and does not require any additional hardware. During the closing and holding phases, the MCU runs in PWM Mode under a PI law that drives the motor until Load reaches a target reference value, whereas the return-to-home phase uses Extended Position Mode with a ramped trajectory.

\subsubsection{Action Parameters}
\label{subsub:Action Parameters}
Three Load reference values (45, 120, and 150) were tested, paired with two PI gain sets (Table~\ref{tab:speed_comparison}). The lower gains (thumb: $K_{p1} = 0.027$, $K_{i1} = 0.32$; index: $K_{p2} = 0.032$, $K_{i2} = 0.64$) were used together with the higher Load reference values (120 and 150), designated as Low-Gain-Load-120 and Low-Gain-Load-150. For a fixed displacement of $190^{\circ}$, the finger completes the motion in $4.88$~s, corresponding to an average angular speed $\omega = 38.93^{\circ}$/s. Conversely, the aggressive movements driven by the higher gains (thumb: $K_{p1} = 0.547$, $K_{i1} = 5.3$; index: $K_{p2} = 1.6$, $K_{i2} = 10.5$), paired with the Load 45 reference value, are designated as High-Gain-Load-45. This configuration completes the same displacement in $2.32$~s, yielding an angular velocity $\omega = 81.90^{\circ}$/s, which represents a speed increase of 110.3\%. The tip-pinch configuration is reproducible with the same controller but was used only as a dexterity demonstration and was excluded from the object classification experiments.

\begin{table}[ht]
\renewcommand{\arraystretch}{1.3}
\caption{Comparison of Pulley Rotational Speeds During Unloaded Finger Motion Across Different Gain and Load Configurations}
\label{tab:speed_comparison}
\centering
\begin{tabular}{lcccc}
\toprule
\textbf{Configuration} & \textbf{Dist. ($^{\circ}$)} & \textbf{Time (s)} & \textbf{Speed ($^{\circ}$/s)} & \textbf{Increase} \\
\midrule
\shortstack{Low-Gain-Load-\\120 \&150} & 0 to 190 & 4.88 & 38.93 & -- \\
High-Gain-Load-45        & 0 to 190 & 2.32 & 81.90 & +110.3\% \\
\bottomrule
\end{tabular}
\end{table}

\subsection{Physical object datasets}
\label{subsec:obj_datasets}
\begin{figure}[!h]
    \centering
    \includegraphics[width=0.48\textwidth]{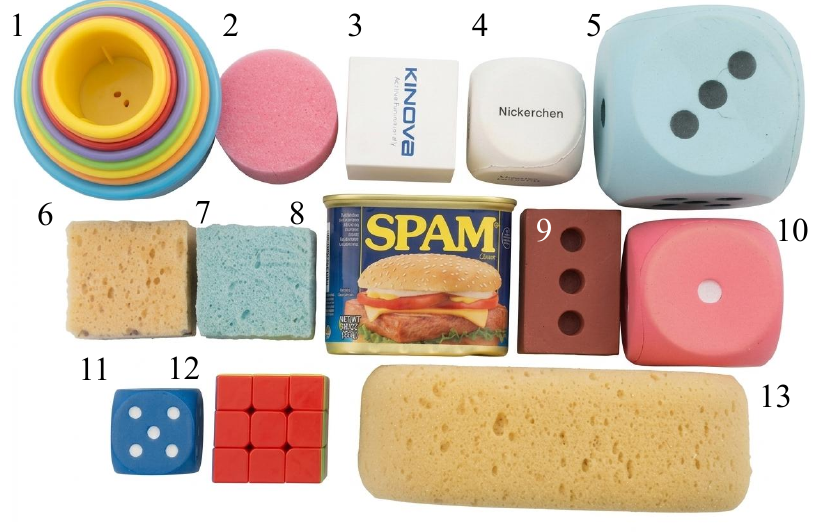}
    \caption{Object sets used in the object recognition experiments. A set of toy cylinders(marked \textbf{1} and \textbf{2}) in the figure, gives systematic variation in size and material. A set of irregular cubes and cuboids(marked \textbf{3-13}) provides a more challenging classification problem.}
    \label{fig:object_sets}
\end{figure}

Two sets of objects were collected, a \textbf{\textit{cylinders}} set and a \textbf{\textit{cubes and cuboids}} set, compiled together in Fig.~\ref{fig:object_sets}. The \textit{cylinders} set is comprised of the set of eight plastic toy cylindrical containers borrowed from the YCB dataset~\cite{calli2015ycb} and supplemented with a pink foam cylinder with equal diameter to one of the containers. This set of objects provides systematic variation in two object properties: size and material.
The rest of the objects form our \textit{cubes and cuboids} set, which provide more non-uniform variability in size and material. Further, many of these objects are deformable in nature, and possess different values of elasticity and how much they can be compressed when a grasping force is applied. Finally, data from empty grasps is also collected.

\subsection{Data collection experiments}
\label{subsec:data_collection}
For each object, data was collected for load values of 45, 120, and 150, described in detail in Table~\ref{tab:speed_comparison}. Each recording consists of placing the open gripper fingers around the object and running a gripper closing routine. Further, for the middle load value of 120, data was collected for different grasping actions. The two different actions explored are labeled as transient grasps for the first grasp and stable grasps for the second grasp. In 'transient' grasps, the object is not perfectly fit within the concavity of the gripper, and thus can move around as the fingers enclose around it. In 'stable' grasps, the object is fit within the gripper and no motion of the object is possible. Data for the two different actions for load value 120 are recorded for 12 of the 20 objects. This is done because the other 8 objects have dimensions that fit tightly into the gripper, and no distinct transient or stable behavior was observed. Overall, the dataset comprises of 4,486 recordings of the STIR hand grasping 20 different objects.

\subsection{Analysis}
\label{subsec:analysis}

\subsubsection{Processing}
\label{subsub:processing}
Seven channels of time-series data, depicted in Fig.~\ref{fig:feedback}, are obtained from the gripper: five from the proposed joint sensors and two from the motor feedback. Each channel's data is passed through the same feature extraction procedure. Fifteen hand-crafted features represent the minimum, mean, and peak values of the signal amplitude, the speed of deformation and resistive force applied by the object on the grasp, the signal-to-noise ratio, and other significant characteristics, as implemented in Hoffmann et al.~\cite{hoffmann2014effect}. Then, each channel's features were concatenated to get a 105-dimensional embedding which describes the interaction between the gripper and the target object. After feature extraction is complete for the entire dataset, the values for each element of the embedding goes through an independent, dataset-wide normalization process. This stops any one feature or channel from dominating the distribution of the data.

\begin{figure*}[htb]
    \centering
    \includegraphics[width=\textwidth]{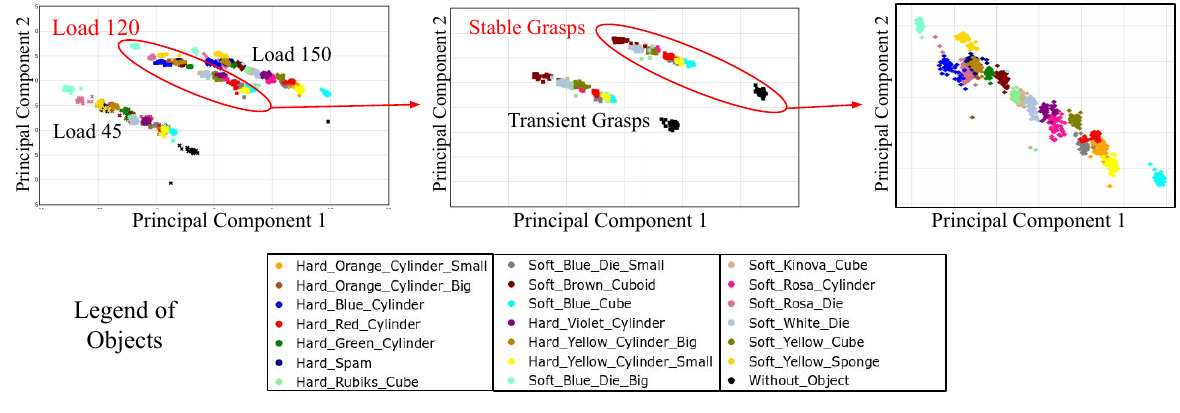}
    \caption{Data visualization using Principal Component Analysis (PCA). Marker colors -- different objects grasped. Black markers -- closing of empty hand (``without object''). (Left) Main clusters formed for different action parameters (High-Gain-Load-45, Low-Gain-Load-120, Low-Gain-Load-150). (Middle) For the Low-Gain-Load-120 setting, additional experiments were conducted to study the effect of the first grasp (``Transient grasps'') and the consecutive grasps (``Stable grasp'') once the object settled in the hand. (Right) Zooming in on the Low-Gain-Load-120 setting and the different objects grasped.}
    \label{fig:clusters}
\end{figure*}

\subsubsection{Unsupervised Analysis}
\label{subsub:unsupervised_analysis}
Next, the data is visualized and analyzed using Principal Component Analysis, a standard statistical and dimension-reduction technique. This allows for the observation of patterns in data, whether object interactions form clusters, or whether any of the embeddings represent physical object characteristics. It also informs the decision of which classification models to deploy. Data where the features and embeddings themselves make the classification easy may only require smaller stochastic or math-based models, whereas more confusing clustering might indicate the need for more complex models, as mentioned in Section~\ref{subsec:haptic_sensing}. The results of this analysis are discussed in depth in Section~\ref{subsec:results_unsup}. 

\subsubsection{Object classification models}
\label{subsub:classifiers}
Based on the results of the unsupervised analysis, the following models/classification methods were chosen for the object classification task:
\begin{itemize}
    \item \textbf{$k$-nearest neighbors ($k$-NN)}:
    An average Euclidean distance metric was used as the classification criterion. $k=3$ was selected empirically by evaluating the performance on the validation set.
    \item \textbf{Support Vector Classifier (SVM)}: 
    The choice for kernel and the value of $C$ are optimized via results on the validation dataset and the speed of training. For this work, the kernel is set to 'linear' and $C=1.0$, because of the achieved high accuracies in most scenarios. The $\gamma$ value is set to 'scale', to scale automatically with the size and variance in the dataset.
    \item \textbf{eXtreme Gradient Boosting}: 
    The training parameters are set via trial-and-error on the validation set.
    \item \textbf{Multi-Layer Perceptron (MLP)}: Our feedforward neural network is a simple three-layer implementation that comprises of a linear expansion layer (105 $\rightarrow$ 256-D) followed by two linear compression layers (256 $\rightarrow$ 64-D), a final compression and softmax layer to the number of object classes. All layers have ReLU activations, and standard cross-entropy loss is used to train the network with an ADAM optimizer and learning rate = 0.001.
\end{itemize}
All methods use our embeddings as input. The first two methods (k-NN and SVM) are implemented via their Scikit-Learn implementations. XGBoost is implemented from the XGBoost Python library. For the MLP classifier, we use PyTorch. For each method, the datasets used were divided into training, validation and testing subsets. Testing and validation subsets were distinct but had the same size. 

\subsubsection{Ablation study}
\label{subsub:ablation}
Later, the entire set of classification experiments is repeated for an ablation study to test the contribution of the motor currents and the proposed, novel joint sensors separately. This is done by masking the features extracted from the joint sensor signals when training the classification models. 

\section{Results}
\label{sec:Results}

\subsection{Unsupervised Analysis}
\label{subsec:results_unsup}
Qualitative, unsupervised analysis of the data was carried out using principal component analysis. This technique is useful for assessing the behavior of data before it is used for training classification models. Further, it can also act as a basis to decide which complexity of models is required. For the collected data, the following observations are made:
\begin{figure}[h]
    \centering
    \includegraphics[width=0.5\textwidth]{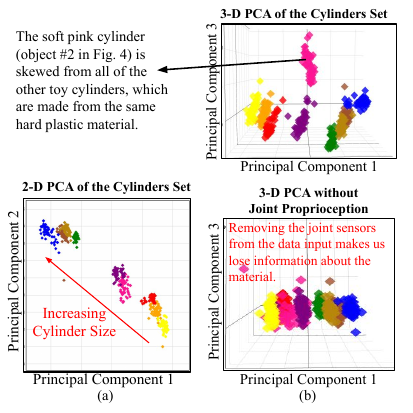}
    \caption{The PCA analysis shows that the bending sensors accurately capture the variations present in object size and object material. See Fig.~\ref{fig:clusters} for the legend of the objects grasped.}
    \label{fig:material_perception}
\end{figure}


\begin{table}[ht]
\renewcommand{\arraystretch}{1.3} 
\caption{Classification results on the unseen testing set using different models and different subsets of data. All values are accuracy percentages on the set of 20 objects compiled in Section~\ref{subsec:obj_datasets}.}
\label{tab:classification_results}
\centering
\begin{tabular}{ccccc}
\toprule
\textbf{Model} & \shortstack{\textbf{High-Gain-}\\\textbf{Load-45} (\%)} & \shortstack{\textbf{Low-Gain-}\\\textbf{Load-120} (\%)}  & \shortstack{\textbf{Low-Gain-}\\\textbf{Load-150} (\%)} & \textbf{All} (\%)\\ 
\midrule
k-NN & 75.2 & 79.8 & 97.8 & 87.8\\
SVM & 95.2 & 96.1 & 99.8 & 98.4\\ 
XGBoost & 98.1 & 97.1 & 99.0 & 99.0\\
MLP & 96.2 & 94.8 & 97.8 & 98.2\\
\bottomrule
\end{tabular}
\end{table}

\begin{enumerate}
    \item In accordance with previous research in~\cite{pliska2024single}, the action configuration and gripper morphology plays an important role in the clustering of data. The largest clusters are based on the 'load' parameter of the compression/grasping action, as described in Section~\ref{subsec:data_collection}. Thus, three clusters are obtained for the three chosen load values: 45, 120, and 150. 
    \item As the experiments also included different actions for the load value of 120, smaller sub-clusters are observed, grouped by the actions carried out during the gripper-object interaction: one set of 'transient-grasps' and a separate set of 'stable-grasps'. Individual object clusters are observed for each action cluster.
    \item[] These first two observations can be seen in Fig.~\ref{fig:clusters}.
    \item In Fig.~\ref{fig:material_perception}, a representation of the systematic variation of the object properties emerges from the data. A 2-dimensional visualization of the \textit{cylinders} object set shows that the feature extraction procedure arranges them in order of their diameters. Further, upon extending the PCA plot to a third dimension, clear separation between cylinders of different stiffness is observed. 
    \item In the ablation study, the same 3-dimensional plot is unable to separate the different objects based on stiffness. This shows that the joint bending sensors are able to perceive object properties that would otherwise have been hidden from grippers using only motor feedback for proprioception.
    \item Lastly, based on the behavior, quality, and quantity of the collected data, the classification models described in Section~\ref{subsub:classifiers} are chosen as suitable candidates.
\end{enumerate}

\subsection{Object Classification}
\label{sec:results_classification}
For every object classification experiment, the training, validation, and test sets are partitioned in the ratio $0.6:0.2:0.2$. Every model is run on the following sets of data: High-Gain-Load-45, Low-Gain-Load-120, Low-Gain-Load-150, and the entire dataset. The accuracies of all the models across all the sets of data including all channels is shown in Table~\ref{tab:classification_results}. Overall, the XGBoost and MLP models performed with the highest accuracy (\textgreater 90\%). The k-NN model performed the worst, but still sufficiently improved upon the random baseline score.

\begin{figure}[h]
    \centering
    \includegraphics[width=0.5\textwidth]{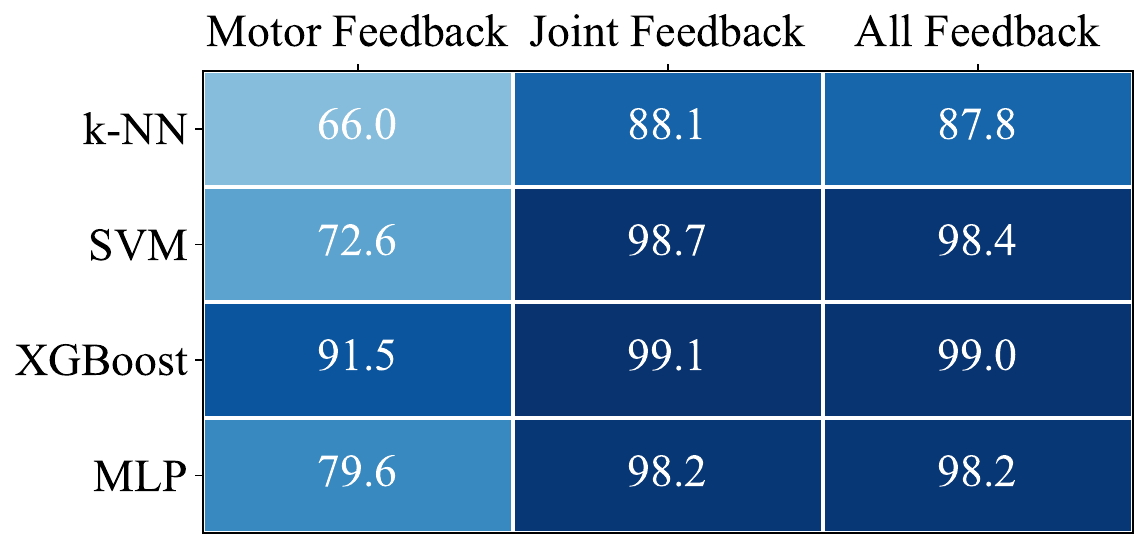}
    \caption{Results of the ablation study. All values are accuracy percentages for the object classification task on the set of 20 objects compiled in Section~\ref{subsec:obj_datasets}.}
    \label{fig:Ablation}
\end{figure}

All the experiments are run three times for an ablation study: once with only the motor feedback, once with only the joint sensor input, and once with all sensory input channels; to analyze the contribution of the novel joint sensors to the object classification task. In addition to the observations made in the previous section about loss of object property perception without the joint sensors, Fig.~\ref{fig:Ablation} shows a quantitative drop in performance without the novel proprioception that has been added to the system.


\subsection{Comparison with other grippers}
\label{subsec:gripper_comparison}
Pliska et al.~\cite{pliska2024single} also analyse different commercially available robotic grippers through the perspective of how effective their proprioception is in helping with the object classification task. In this previous work, the four chosen grippers are the qb SoftHand(qb SH), the Robotiq 2F-85(2F-85), the OnRobot RG6(RG6), and the Barrett Hand(BH). In addition to a k-NN classifier and an SVM classifier, a Long Short-Term Memory(LSTM) model was used directly on the raw time-series signal without any feature extraction process. To add the STIR Hand to this comparison, an additional LSTM model was trained on a subset of the collected data. This subset included all recordings of nine objects from the \textit{cubes and cuboids} set, the same objects used in the previous work. The performance of the STIR hand is presented in Table~\ref{tab:gripper_comparison} in comparison with the performance of the other grippers, borrowed from the previous work. Of particular interest is the direct comparison to another compliant gripper, the qb SoftHand, which does not possess any joint proprioception. The STIR hand outperforms the SoftHand on every model, further highlighting the advantage provided by the integrated joint sensors. 

\begin{table}[ht]
\renewcommand{\arraystretch}{1.3}
\caption{Object classification accuracy comparison with other grippers. All numbers are accuracy percentages on the same 9 objects from the \textbf{\textit{cubes and cuboids}} dataset.}
\label{tab:gripper_comparison}
\centering
\begin{tabular}{c|cc|ccc}
\toprule
\multirow{2}{*}{\textbf{Model}} & \multicolumn{2}{c|}{Compliant Grippers} & \multicolumn{3}{c}{Rigid Grippers}\\
\cmidrule(lr){2-3} \cmidrule(lr){4-6}
 & \textbf{STIR} & \textbf{qb SH} & \textbf{2F-85} & \textbf{RG6} & \textbf{BH} \\
 \midrule
 K-NN (feat.) & \textbf{89.6} & 59.8 & 97.4 & 80.4 & 96.7 \\
 SVM & \textbf{98.4} & 70.7 & 99.1 & 97.0 & 100.0 \\
 LSTM & \textbf{85.6} & 74.3 & 99.3 & 97.8 & 99.1 \\
\bottomrule
\end{tabular}
\end{table}

\section{Conclusion and Future Work}
\label{sec:conclusion}

Our experiments confirm that integrating proprioceptive sensors into the hand's structural morphology effectively boost perception while enabling the passive compliance typical of soft robotic systems. This low-cost, embedded kinesthetic sensing design optimize grasp state estimation without requiring external vision devices. 

Several key areas remain for future investigation. First, miniaturizing the STIR Hand  and electronics is essential to facilitate integration with mobile manipulation platforms, make it self-contained, and expand its capabilities to grasp smaller, heavier, or asymmetric and pyramidal geometries. Optimizing the fingertip geometry and the related control for a precise tip-pinch grasp is essential for effective object retrieval from flat surfaces. The system must undergo thorough evaluation under less structured environments and tasks like varied approach angles, transient pressure fluctuations, and in-motion manipulation via robotic arms or human teleoperation. 
While the current morphology performs exceptionally well, integrating fingertip tactile sensors could further enhance object discrimination accuracy and enable slip detection via friction monitoring, allowing the system to dynamically adjust grasping power as needed.

This hardware foundation will allow the framework to transition from offline state estimation to true real-time, closed-loop grasp control using joint proprioception. Consequently, the hand will be capable of automatically modulating its grasping forces during physical interaction. Looking ahead, the integrated algorithmic frameworks will evolve beyond static object identification toward continuous property estimation, leveraging real-time sensor trends to dynamically predict object size and flexibility.

\section{Acknowledgements}
We would like to thank Bedřich Himmel for consultations and assistance with the experimental setup. Also, thanks to Federico Carli and Federico Micheletti for assisting with the custom PCBs.

\bibliographystyle{IEEEtran}
\bibliography{STIR-Hand}

\end{document}